\documentclass[10pt,twocolumn,letterpaper]{article}

\usepackage[pagenumbers]{cvpr} 

\definecolor{cvprblue}{rgb}{0.21,0.49,0.74}
\usepackage[pagebackref,breaklinks,colorlinks,allcolors=cvprblue]{hyperref}
\usepackage{array}
\usepackage{tabularx}
\usepackage{indentfirst}
\usepackage{colortbl}
\usepackage[misc]{ifsym} 

\def\paperID{10716} 
\def\confName{CVPR}
\def\confYear{2025}
\title{DefMamba: Deformable Visual State Space Model}

\author{ Leiye Liu$^{1}$ \hspace{.2cm} Miao Zhang$^{1,}$\footnotemark[1] \hspace{.2cm} Jihao Yin$^{1}$ \hspace{.2cm} Tingwei Liu${^1}$ \hspace{.2cm} Wei Ji${^2}$ \hspace{.2cm} Yongri Piao$^{1,}$\footnotemark[1] \hspace{.2cm} Huchuan Lu${^1}$ \\
 $^1$Dalian University of Technology  \hspace{.5cm}   $^2$Yale University \\
{\tt\small\ leiyeliu@mail.dlut.edu.cn, wei.ji@yale.edu, yrpiao@dlut.edu.cn}
}

\begin{document}
\maketitle
\renewcommand{\thefootnote}{\fnsymbol{footnote}}
\footnotetext[1]{Corresponding authors.}
\begin{abstract}
Recently, state space models (SSM), particularly Mamba, have attracted significant attention from scholars due to their ability to effectively balance computational efficiency and performance. However, most existing visual Mamba methods flatten images into 1D sequences using predefined scan orders, which results the model being less capable of utilizing the spatial structural information of the image during the feature extraction process. To address this issue, we proposed a novel visual foundation model called DefMamba. This model includes a multi-scale backbone structure and deformable mamba (DM) blocks, which dynamically adjust the scanning path to prioritize important information, thus enhancing the capture and processing of relevant input features. By combining a deformable scanning (DS) strategy, this model significantly improves its ability to learn image structures and detects changes in object details. Numerous experiments have shown that DefMamba achieves state-of-the-art performance in various visual tasks, including image classification, object detection, instance segmentation, and semantic segmentation. 
The code is open source on \href{https://github.com/leiyeliu/DefMamba}{DefMamba} .
\end{abstract}    
\section{Introduction}
\label{sec:intro}

Most existing visual foundation models primarily rely on convolutional neural networks (CNNs) \cite{convnext, regnety, effnet} and Transformer architectures \cite{vit, swin, deit}. However, CNNs are constrained by their sliding window structure, which limits the receptive field and significantly impedes global information aggregation across the input data. In contrast, Transformers excel in global information aggregation due to their attention mechanism, but their high computational complexity poses a challenge in achieving a balance between efficiency and performance. State space models (SSMs) \cite{s4} provide a potential solution to this trade-off. SSMs aggregate previous features through a hidden state matrix to update current features, thereby reducing the computational complexity to a linear relationship with the sequence length. Although SSMs process sequences in a recurrent manner, SSMs can perform calculations on sequences in parallel after simplification. Despite these advantages, SSMs struggle to capture long-range dependencies due to the lack of content-aware perception in the state matrix update process.

\begin{figure}
    \centering
    \includegraphics[width=\columnwidth]{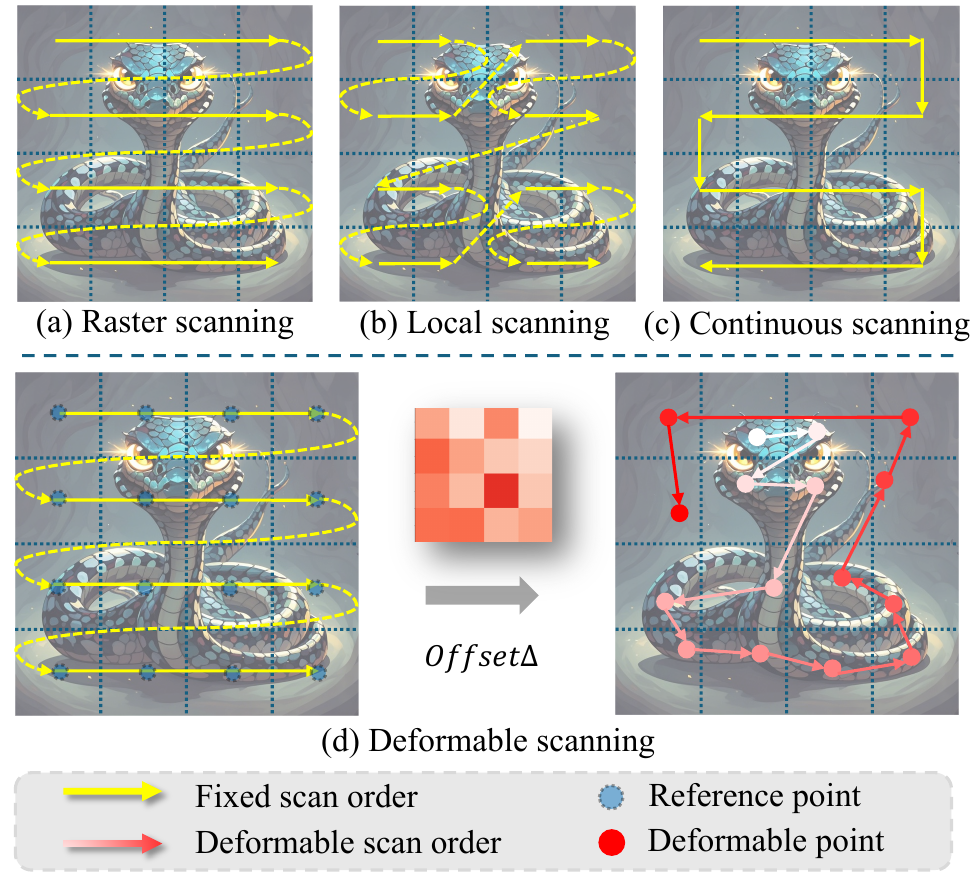}
    \caption{Comparison of deformable scanning and other scanning methods. (a) Raster scanning \cite{vim, vmamba}, (b) Local scanning \cite{localmamba}, (c) Continuous scanning \cite{plainmamba}, (d) Our designed deformable scanning. The blue dots represent the reference points, and the red dots represent the deformable points. The yellow arrows represent the fixed scan order, and the red gradient arrows represent the deformable scan order. Our method exhibits an enhanced capacity to accurately capture the structural characteristics of objects, thereby enabling the development of a more refined scanning approach.}
    \label{fig:motivation}
\end{figure}

Recently, Mamba \cite{mamba} proposes an improved selection mechanism designed to optimize the training process of SSMs. This innovative mechanism introduces content awareness into the feature extraction pipeline, expands the effective receptive field, and achieves remarkable performance enhancements in various NLP tasks. Consequently, numerous studies have attempted to extend this approach to the broader field of computer vision \cite{shapemamba, swinumamba, pathmamba, segmamba, mim, rsmamba, spectralmamba, Samba, learning, DGFamba}. The principal challenge in this endeavor is how to effectively map 2D image feature maps into 1D sequences, without losing essential information. Most existing methods employ predefined strategies for mapping, such as raster scanning \cite{vim, vmamba}, local scanning \cite{localmamba}, and continuous scanning \cite{plainmamba}. However, as illustrated in Figure \ref{fig:motivation}, these methods all rely on a fixed scanning path. This results in adjacent tokens no longer being adjacent after flattening. Thereby, they neglect the inherent spatial structure of the image, leading to the loss of structural information. To address this issue, QuadMamba \cite{quadmamba} determines the scanning window size based on the amount of information contained in different areas of the image. However, the scanning order in each window is fixed. This leads to an incomplete solution of the aforementioned issues. GrootV \cite{grootvl} adaptively constructs a tree topology based on input features and subsequently extracts features from this topology. Nevertheless, it employs only adjacent features in constructing the topology and distributes attention uniformly across the patch.  The aforementioned methods are either based on a fixed scanning order, leading to the loss of structural information, or treat the information in the perception area equally, resulting in insensitivity to variations in object details.

To solve this problem, we proposed a novel framework called DefMamba inspired by deformable mechanisms \cite{dcn, dat}. However, intuitively applying deformable mechanisms to SSMs still causes structural information loss and increases computational complexity. Therefore, we proposed a deformable state space model and a deformable scanning strategy (DS) to prioritize deformable tokens based on essential information and slide reference points towards the important area. This approach enables the SSMs to capture and process relevant features related to the input more effectively. Specifically, we shifted the reference point by the generated offset from a fixed position to an adjustable one that provided more useful information, thereby facilitating the awareness of changes in object details. On the other hand, we also dynamically adjusted the scanning order by the offset vector for obtaining a structure-aware sequence. In this way, our framework adaptively perceives the variations in object details to find the most suitable feature points, and determines the optimal scanning order consistent with the object structure based on the input image features. 

We conducted extensive experiments to validate the effectiveness of DefMamba across multiple visual benchmarks, including image classification on ImageNet \cite{imagenet}, object detection and instance segmentation on COCO \cite{coco}, and semantic segmentation on ADE20K \cite{ade20k}. These results demonstrate that our method  outperforms existing SSMs based approaches on all benchmarks and remains competitive with CNN and transformer-based methods.
\section{Related Work}
\label{sec:relatework}

\subsection{Mamba for Visual Applications}
Numerous studies have successfully integrated Mamba \cite{mamba} into visual tasks \cite{vim,vmamba,plainmamba,efficientvmamba,localmamba,MSV,groupmamba,grootvl,quadmamba}, demonstrating preliminary achievements. ViM \cite{vim} introduces a bidirectional scanning approach to transform 2D image into 1D sequences, which are then fed into SSM for global context modeling, marking the first integration of Mamba into visual tasks. VMamba \cite{vmamba} employs a four-way scanning algorithm to convert 2D image into 1D sequences. PlainMamba \cite{plainmamba} modifies the scanning method from raster to continuous, preserving the spatial dependencies of the image. MSVMamba \cite{MSV} downsamples the sequence based on four scans to reduce computational redundancy and mitigates the issue of information loss. GrootV \cite{grootvl} constructs a minimum spanning tree on a four-way plane graph using the differences between adjacent features, dynamically adjusting the scanning order according to different inputs. QuadMamba \cite{quadmamba} adaptively adjusts the window granularity during the scanning process based on the information content of the image to better aggregate local information. While GrootV and QuadMamba can adapt their scanning methods based on input data, GrootV only considers the relationships between adjacent elements when generating the minimum spanning tree, neglecting global information. On the other hand, QuadMamba still relies on predefined scanning methods and does not achieve true dynamic scanning. In contrast, our DefMamba introduces a content-aware deformable scanning strategy that allows the network to dynamically learn the scanning order and the reference points position.

\begin{figure*}
    \centering
    \includegraphics[width=\textwidth]{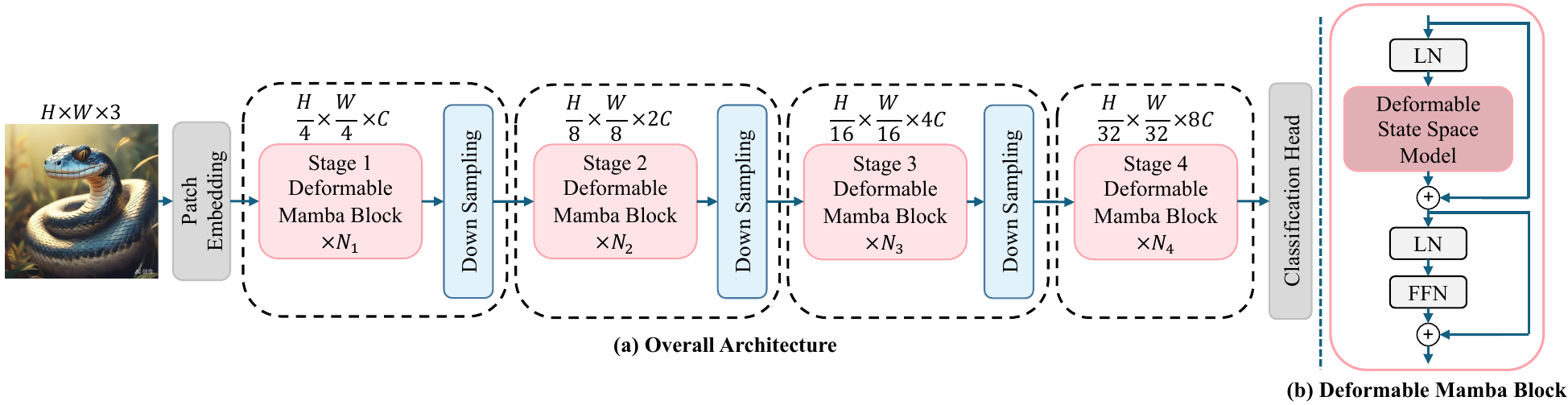}
    \caption{Overview of DefMamba. (a) depicts the overall architecture of our network. (b) illustrates the structure of the deformable Mamba block. LN means LayerNorm and FFN is a feed-forward network.}
    \label{fig:overall}
\end{figure*}

\subsection{Deformable CNNs and Attention Mechanism}
Deformable convolution \cite{dcn,dcnv2,dcnv3,dcnv4} employs a convolution kernel that adapts to geometric variations in the input feature map, thereby overcoming the limitations of traditional convolution, which performs poorly when dealing with complex targets. Recently, the deformable mechanism has been extended to visual transformers \cite{vit} to enhance their ability to capture local features and adapt to geometric variations. DPT \cite{dpt} proposes an adaptive patch embedding method that dynamically adjusts the position and size of patches while preserving their semantic information. PS-ViT \cite{psvit} introduces a progressive sampling module prior to ViT \cite{vit}, which iteratively identifies the most suitable deformable point positions for the current image. DAT \cite{dat} integrates the deformable mechanism with the self-attention mechanism in Transformers for the first time, incorporating deformable attention into the visual backbone. This approach learns a set of features corresponding to global keypoints and adapts to spatial variations. Previous approaches have explored various ways to effectively incorporate deformable mechanism into transformer architectures. With the recent popularity of Mamba \cite{mamba}, we attempted to introduce deformable mechanism into Mamba. However, directly applying these mechanism has resulted in issues like the loss of structural information and the necessity for additional modules. In this context, our designed DS strategy stands out by effectively prioritizing deformable tokens while directing reference points toward key areas.

\section{Method}
\label{sec:method}

In this section, we first summarized the SSMs in Section 3.1. Then, in Section 3.2, we described the overall structure of the proposed network. Section 3.3 introduces a Deformable State Space Model (DSSM). Finally, Section 3.4 presents the designed model configurations across multiple scales.

\subsection{Preliminaries}
SSMs, including notable implementations like S4 \cite{s4} and Mamba \cite{mamba}, are structured sequence architectures that combine elements of recurrent neural networks (RNNs) and CNNs, enabling linear or near-linear scaling with respect to sequence length. These models, derived from continuous systems, define a 1D function-to-function map for an input \(u(t) \in \mathbb{R}^L\) to an output \(y(t) \in \mathbb{R}^L\) through a hidden state \(h(t) \in \mathbb{R}^N\). Where $t$ represents time. More formally, SSMs are characterized by the continuous-time Ordinary Differential Equation (ODE) \cite{mamba} presented:

\vspace{-2pt}
\begin{small}
\begin{equation}
\begin{aligned}
    h'(t) &= \mathbf{A}h(t) + \mathbf{B}u(t), \\
    y(t) &= \hat{C}h(t),
\end{aligned}
\end{equation}
\end{small}

\noindent where \(h(t)\) is the current hidden state. \(h'(t)\) is the updated hidden state. \(u(t)\) is the current input. \(y(t)\) is the output. \(\mathbf{A} \in \mathbb{R}^{N \times N}\) is SSM's evolution matrix, and \(\mathbf{B} \in \mathbb{R}^{N \times 1}\), \(\hat{C} \in \mathbb{R}^{1 \times N}\) are the input and output projection matrices, respectively.

To enable the application of SSMs in sequence modeling tasks within deep learning, they must be discretized, transforming the SSM from a continuous-time function-to-function map into a discrete-time sequence-to-sequence map. S4 \cite{s4} and Mamba \cite{mamba} are examples of discrete adaptations of the continuous system, incorporating a timescale parameter $\Delta$ to convert the continuous parameters $\mathbf{A}$, $\mathbf{B}$ into their discrete counterparts $\bar{\mathbf{A}}$, $\bar{\mathbf{B}}$. This discretization is typically achieved using the Zero-Order Hold (ZOH) \cite{mamba} method:

\vspace{-5pt}
\begin{small}
\begin{equation}
\begin{aligned}
    \bar{\mathbf{A}} &= \exp(\Delta \mathbf{A}), \\
    \bar{\mathbf{B}} &= (\Delta \mathbf{A})^{-1} (\exp(\Delta \mathbf{A}) - \mathbf{I}) \cdot \Delta \mathbf{B}, \\
    h_t &= \bar{\mathbf{A}} h_{t-1} + \bar{\mathbf{B}} u_t, \\
    y_t &= \mathbf{C} h_t.
\end{aligned}
\end{equation}
\end{small}

While both S4 \cite{s4} and Mamba \cite{mamba} employ a similar discretization process as outlined in Equation 2, Mamba distinguishes itself from S4 by conditioning the parameters $\Delta \in \mathbb{R}^{b \times L \times D}$, $\mathbf{B} \in \mathbb{R}^{b \times L \times N}$, and $\mathbf{C} \in \mathbb{R}^{b \times L \times N}$ on the input $u \in \mathbb{R}^{b \times L \times D}$ through the S6 Selective Scan Mechanism. Here, $b$ represents the batch size, $L$ denotes the sequence length, and $D$ signifies the feature dimension.

\begin{figure*}
    \centering
    \includegraphics[width=\textwidth]{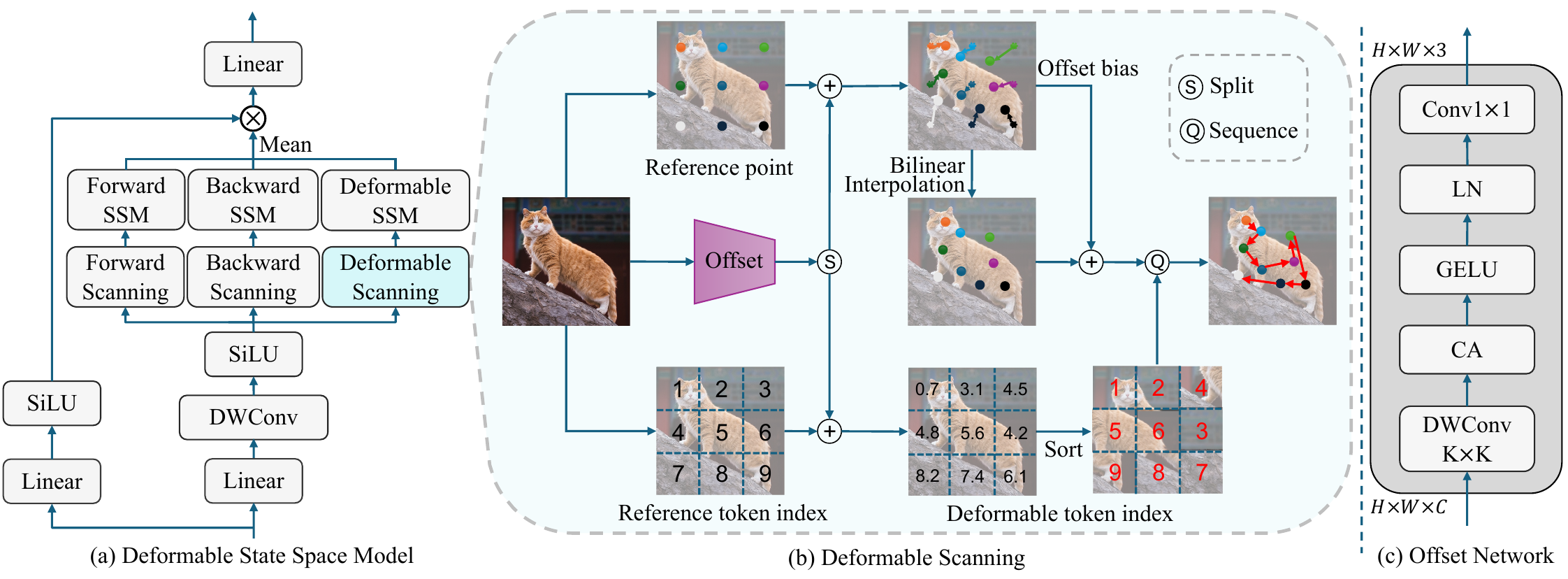}
    \caption{Illustration of Deformable State Space Model. (a) illustrates the processing flow of the deformable state space model for feature extraction. (b) depicts the processing flow of the deformable scan. The upper part primarily shifts the feature points to enable the model to focus on more salient regions, while the lower part shifts the token positions to facilitate the discovery of a scanning order that is better suited to the current input. To clearly illustrate the process, only nine points are depicted in the figure, however, the actual processing involves a greater number of points. (c) presents the detailed structure of the offset network.}
    \label{fig:dem}
\end{figure*}

\subsection{Overall Model Architecture}
DefMamba uses a common multi-scale backbone structure similar to many CNNs \cite{regnety, convnext} and Transformers \cite{swin}. As illustrated in Figure \ref{fig:overall}(a), the image $I \in \mathbb{R}^{H \times W \times 3}$ is first divided into patches through a patch embedding layer, which produces a 2D feature map with spatial dimensions $H/4 \times W/4$ and channel dimensions $C$. Subsequently, multiple network stages are used to create hierarchical representations of dimensions $H/8 \times W/8 \times 2C$, $H/16 \times W/16 \times 4C$, and $H/32 \times W/32 \times 8C$. Each stage consists of a stack of Deformable Mamba (DM) blocks followed by a downsampling layer (except the fourth stage). Finally, the features are average-pooled and sent to the classification head to obtain the prediction results. In particular, we followed \cite{grootvl} and used overlapping forms in the patch embedding layer and the downsampling layers. For specific structural details, please refer to the Appendix.

Different from the Mamba structure used in language models \cite{mamba}, the DM block follows the popular structure of the Transformer block \cite{vit, swin}, which consists of two Layer Norm (LN) layers, an FFN, a DSSM (to be introduced in the following subsection) and residual connections, as shown in the Figure \ref{fig:overall}(b).

\subsection{Deformable State Space Model}
The overall architecture of the deformable state space model is presented in Figure \ref{fig:dem}(a). Inspired by \cite{vmamba, MSV, grootvl}, we employed depthwise convolution to replace the original 1D convolution in the vision mamba block \cite{vim} and incorporated a deformable branch comprising a deformable scanning and a deformable SSM (DSSM). We maintained the standard forward and backward branches to ensure stable model convergence, as our method introduces more spatial token jumps compared to previous scanning methods, potentially complicating model training and learning. Subsequent experiments validate this (Table \ref{tab:alb1}).

\noindent \textbf{Deformable Scanning}. Given the issue of mutual interference between multiple deformable points, we constrained the offset of the deformable points in a certain range. This ensures the relationship between deformable points and reference points remains invariant, allowing us to approximate the relationship after deformation using the relationship before deformation. Furthermore, considering the simplicity of the computation, we employed a parallel approach to simultaneously adjust the reference points and scanning order, thereby reducing the computational burden. The specific structure is illustrated in the Figure \ref{fig:dem}(b).

\begin{table*}[!t]
  \centering
  \begin{minipage}[t]{0.49\linewidth} 
    \centering
    \resizebox{\linewidth}{!}{ 
\begin{tabular}{>{\raggedright\arraybackslash}m{4cm} >{\centering\arraybackslash}m{0.8cm} >{\centering\arraybackslash}m{1cm} >{\centering\arraybackslash}m{1.2cm} >{\centering\arraybackslash}m{0.8cm}}
\toprule[2pt]
\textbf{Method} & \textbf{Type} & \textbf{\#Param.} & \textbf{\#FLOPs} & \shortstack{\textbf{Top-1} \\ \textbf{Acc.}} \\
\midrule[1pt]
RegNetY-800M \cite{regnety} & C & 6M & 0.8G & 76.3 \\
GhostNet 1.3$\times$ \cite{ghostnet} & C & 7M & 0.2G & 75.7 \\
DeiT-Ti \cite{deit} & T & 6M & 1.3G & 72.2 \\
ViM-T \cite{vim} & S & 7M & 1.5G & 76.1 \\
EffVMamba-T \cite{efficientvmamba} & S & 6M & 0.8G & 76.5 \\
LocalVim-T \cite{localmamba} & S & 8M & 1.5G & 76.2 \\
PlainMamba-L1 \cite{plainmamba} & S & 7M & 3.0G & 77.9 \\
MSVMamba-N \cite{MSV} & S & 7M & 0.9G & 77.3 \\
QuadMamba-Li \cite{quadmamba} & S & 5M & 0.8G & 74.2 \\
\rowcolor{gray!20}
\textbf{DefMamba-T} & S & 8M & 1.2G & \textbf{78.6} \\
\midrule[1pt]
RegNetY-4G \cite{regnety} & C & 21M & 4.0G & 80.0 \\
ConvNeXt-T \cite{convnext} & C & 29M & 4.5G & 82.1 \\
Conv2Former-T \cite{conv2former} & C & 27M & 4.4G & 83.2 \\
DeiT-S \cite{deit} & T & 22M & 4.6G & 79.8 \\
Swin-T \cite{swin} & T & 29M & 4.5G & 81.3 \\
CoAtNet-0 \cite{coatnet} & T & 25M & 4.0G & 81.6 \\
CrossFormer-S \cite{crossformer} & T & 31M & 4.9G & 82.5 \\
ViM-S \cite{vim} & S & 26M & 5.1G & 81.0 \\
VMamba-T \cite{vmamba} & S & 22M & 5.6G & 82.2 \\
\bottomrule[2pt]
\end{tabular}

    }
  \end{minipage}
  \hspace{-0mm}
  \begin{minipage}[t]{0.49\linewidth} 
    \centering
    \resizebox{\linewidth}{!}{ 
      \begin{tabular}{>{\raggedright\arraybackslash}m{4cm} >{\centering\arraybackslash}m{0.8cm} >{\centering\arraybackslash}m{1cm} >{\centering\arraybackslash}m{1.2cm} >{\centering\arraybackslash}m{0.8cm}}
\toprule[2pt]
\textbf{Method} & \textbf{Type} & \textbf{\#Param.} & \textbf{\#FLOPs} & \shortstack{\textbf{Top-1} \\ \textbf{Acc.}} \\
\midrule[1pt]
LocalVim-S \cite{localmamba} & S & 28M & 4.8G & 81.2 \\
LocalVMamba-T \cite{localmamba} & S & 26M & 5.7G & 82.7 \\
EffVMamba-B \cite{efficientvmamba} & S & 33M & 4.6G & 83.0 \\
MSVMamba-T \cite{MSV} & S & 33M & 4.6G & 82.8 \\
PlainMamba-L2 \cite{plainmamba} & S & 25M & 8.1G & 81.6 \\
GrootV-T \cite{grootvl} & S & 30M & 4.8G & 83.4 \\
QuadMamba-S \cite{quadmamba} & S & 31M & 5.5G & 82.4 \\
\rowcolor{gray!20}
\textbf{DefMamba-S} & S & 32M & 4.8G & \textbf{83.5} \\
\midrule[1pt]
ConvNeXt-S \cite{convnext} & C & 50M & 8.7G & 83.1 \\
Conv2Former \cite{conv2former} & C & 50M & 8.7G & 84.1 \\
Swin-S \cite{swin} & T & 50M & 8.7G & 83.0 \\
CoAtNet-1 \cite{coatnet} & T & 42M & 8.0G & 83.3 \\
CrossFormer-B \cite{crossformer} & T & 52M & 9.2G & 83.4 \\
VMamba-S \cite{vmamba} & S & 50M & 8.7G & 83.6 \\
LocalVMamba-B \cite{localmamba} & S & 50M & 11.4G & 83.7 \\
PlainMamba-L3 \cite{plainmamba} & S & 50M & 14.4G & 82.3 \\
GrootV-S \cite{grootvl} & S & 51M & 8.5G & 84.2 \\
QuadMamba-B \cite{quadmamba} & S & 50M & 9.3G & 83.8 \\
\rowcolor{gray!20}
\textbf{DefMamba-B} & S & 51M & 8.5G & \textbf{84.2} \\
\bottomrule[2pt]
\end{tabular}
    }
  \end{minipage}
  \caption{Image classification performance on ImageNet-1K validation set. C, T and S indicate the model type of CNNs, Transformer and SSM. The best results are shown in \textbf{bold} font.}
  \label{tab:imagenet}
\end{table*}

Given an input feature $ x \in \mathbb{R}^{H \times W \times C} $, where $ C $ denotes the channel dimension and $ H, W $ represent the spatial resolution. We first generated the offset $ o \in \mathbb{R}^{H \times W \times 3} $ using a subnetwork that utilizes $x$ to output the offset values $ o $ for the reference point and the reference token. 

We initially implemented the subnetwork as depicted in Figure \ref{fig:dem}(c). The input features are first processed through a $ K\times K $ depthwise convolution to capture local features. Subsequently, GELU, Layer Normalization (LN), and a 1$\times$1 convolution are employed to derive the offset values, which encompass three dimensions in total. The first two dimensions represent the offset of the reference point in two-dimensional space, while the third dimension signifies the offset of the reference token index within the entire patch. During our experiments, we observed that the offset of the token necessitates global perception of the features within the patch, which cannot be achieved solely through convolution. In light of this, and considering the previous method's findings \cite{groupmamba, MSV} that mamba has redundancy in the channel dimension, we incorporated a Channel Attention (CA) mechanism \cite{senet} following the depthwise convolution layer. This mechanism mitigates channel redundancy and facilitates the integration of global contextual information. Notably, following the configuration of DAT \cite{dat}, we set $ K $ to $\left [ 9, 7, 5, 3 \right ]$ across four stages and omitted the bias in the 1$\times$1 convolution.

To stabilize the training process, we employed the $tanh$ function to mitigate the impact of extreme values in $ o $, where $ \hat{o} = \tanh(o) $. Subsequently, we split $\hat {o} $ to 2 parts along the channel dimension, one with 2 channels and one with 1 channel, to obtain the point offset $\Delta p $ and token index offset $\Delta t $. As previously mentioned, we needed to further constrain $ \Delta p $, in order to stabilize training and simplify the structure. We divided the horizontal and vertical dimensions of $ \Delta p $ by $ W $ and $ H $, respectively, thereby limiting the offset to the range of a single token. The detailed process is outlined as follows:

\vspace{-10pt}
\begin{small}
\begin{equation}
\begin{aligned}
    \Delta p, \Delta t &= Split( \tanh( \textit{Offset}(x)), \text{dims}=[2, 1]), \\
    \hat{\Delta p} &= \textit{Norm}(\Delta p). \\
\end{aligned}
\end{equation}
\end{small}
\vspace{-8pt}

Then, we sent the point offset $ \hat{\Delta p} $, token index offset $ \Delta t $, and input feature $ x $ to the point offset branch and index branch, to obtain the final output, respectively.

\noindent \textbf{Point Offset}. To obtain feature representations that are more sensitive to changes in objects, we dynamically adapted the network's reference points to deformable points that contain more relevant information based on the input. Firstly, we generated reference points $ p \in \mathbb{R}^{H \times W \times 2} $. The values in $ p $ correspond to the two-dimensional coordinates of points ranging from (0, 0) to ($H-1$, $W-1$). To simplify network calculations, we normalized $ p $ from its original range to [-1, 1], where [-1, 1] represents the point in the upper left corner and [1, 1] represents the point in the lower right corner. We then added the reference point $ p $ and the offset to obtain the deformable point $ \hat{p} = p + \Delta p $. Since the offset $ \hat{p} $ contains a decimal part, it cannot be used directly. Therefore, we used bilinear interpolation to extract features at the spatial position corresponding to the offset point $ \hat{p} $ from the input $ x $.

When our model performs point offset, the features will move in space, potentially causing the position encoding added in the initial stage to become ineffective and thereby decreasing model performance. To address this, we designed an offset bias based on the relative position encoding in the Swin Transformer. Specifically, given the feature map of size $ H \times W $, the relative coordinate displacement of points lies within the range of $\left [ -H, H \right ]$ and $\left [ -W, W \right ]$ in two dimensions, respectively. Therefore, we set a learnable relative offset bias matrix $ R \in \mathbb{R}^{\left (2H - 1 \right ) \times \left (2W - 1 \right )} $. However, considering that such a matrix would result in a significant increase in parameter count, we performed a downsampling operation on this matrix to obtain $ R \in \mathbb{R}^{H \times W} $. At the same time, we divided the point displacement by 2 to accommodate this change. Then we used the point displacement to calculate the corresponding compensation by interpolating on $ R $. Finally, this compensation is added to the interpolated features. The specific process is as follows:

\vspace{-5pt}
\begin{small}
\begin{equation}
\begin{aligned}
    \hat{p} &= p + \Delta p, \\
    \hat{x} &= \phi(x, \hat{p}) + \phi(R, \hat{p}), \\
\end{aligned}
\end{equation}  
\end{small}
\vspace{-5pt}

\noindent where $\phi(x, \hat{p})$ represents the use of bilinear interpolation function to extract the feature corresponding to position $\hat{p}$ on $x$.

\noindent \textbf{Index Offset}. We modified the scanning order to enable the model to perceive the structure of the input object effectively by varying the reference token index and the deformable token index. We initially generated the reference token index $ t_{r} \in \mathbb{R}^{N \times 1} $, where $ N = H \times W $. The values in $ t $ indicate the token positions within the current patch, ranging from 0 to $ N-1 $. To simplify network computations, we normalized $ t_{r} $ from its original range to [-1, 1]. This enables us to compute the deformable token index $ t_{d} = t_{r} + \Delta t $. Since the derived $ t_{d} $ contain decimal components, they cannot be utilized directly. Consequently, for $ t_{d} $, we applied a sorting algorithm to determine the indices post-offset based on the magnitude of their values. Finally, we transformed the offset features $ \hat{x} $ into a 1D sequence according to these indices, thereby obtaining a content-adaptive image feature sequence. It is important to note that the sorting algorithm truncates gradients, rendering the network untrainable. To address this, we averaged the gradients of the final image sequence across the dimension and replicate them to $ \Delta t $ to approximate the gradient of the scanning order offset.

\begin{table}
    \belowrulesep=0pt
    \aboverulesep=0pt
    \centering
    \resizebox{\linewidth}{!}{\setlength{\arrayrulewidth}{1pt} 
\begin{tabular}{l|c|ccc|ccc}
\toprule[2pt]
\multicolumn{8}{c}{\textbf{Mask R-CNN 1$\times$ schedule}} \\
\midrule[1pt]
\textbf{Backbone} & \textbf{\#FLOPs} & \textbf{AP\textsuperscript{b}} & \textbf{AP\textsuperscript{b}\textsubscript{50}} & \textbf{AP\textsuperscript{b}\textsubscript{75}} & \textbf{AP\textsuperscript{m}} & \textbf{AP\textsuperscript{m}\textsubscript{50}} & \textbf{AP\textsuperscript{m}\textsubscript{75}} \\
\midrule[1pt]
ResNet-50 \cite{resnet} & 260G & 38.2 & 58.8 & 41.4 & 34.7 & 55.7 & 37.2 \\
Swin-T \cite{swin} & 267G & 42.7 & 65.2 & 46.8 & 39.3 & 62.2 & 42.2  \\
ConvNeXt-T \cite{convnext} & 262G & 44.2 & 66.6 & 48.3 & 40.1 & 63.3 & 42.8  \\
PVTv2-B2 \cite{pvtv2} & 309G & 45.3 & 67.1 & 49.4 & 41.2 & 64.2 & 44.4  \\
VMamba-T \cite{vmamba} & 286G & 47.4 & 69.5 & 52.0 & 42.7 & 66.3 & 46.0  \\
LocalVMamba-T \cite{localmamba} & 291G & 46.7 & 68.7 & 50.8 & 42.2 & 65.7 & 45.5  \\
QuadMamba-S \cite{quadmamba} & 301G & 46.7 & 69.0 & 51.3 & 42.4 & 65.9 & 45.6 \\
MSVMamba-T \cite{MSV} & 252G & 46.9 & 68.8 & 51.4 & 42.2 & 65.6 & 45.4 \\
GrootV-T \cite{grootvl} & 265G & 47.0 & 69.4 & 51.5 & 42.7 & 66.4 & 46.0 \\
\rowcolor{gray!20}
\textbf{DefMamba-S} & 268G & 47.5 & 69.6 & 51.7 & 42.8 & 66.3 & 46.2 \\
\bottomrule[2pt]
\end{tabular}} 
    \caption{Object detection and instance segmentation performance on MSCOCO 2017 val set. All using Mask R-CNN framework. AP\textsuperscript{b} and AP\textsuperscript{m} indicate the mean Average Precision (mAP) of detection and segmentation, respectively}
    \label{tab:coco}
\end{table}


\section{Experiments}
\label{sec:exper}

\subsection{Image Classification}

\noindent \textbf{Settings.} The image classification experiments are conducted using the ImageNet-1K \cite{imagenet} dataset, which comprises over 1.28 million training images and 50,000 validation images across 1,000 categories. Our training setup closely follows the methodology of previous practices \cite{grootvl, MSV, vmamba}, incorporating various data augmentations such as random cropping, random horizontal flipping, label-smoothing regularization, mixup, autoaugment, and random erasing. The models are trained for 300 epochs using the AdamW \cite{adamw} optimizer with a cosine decay learning rate scheduler, including a 20-epoch warm-up period. The total batch size is set to 1,024, with the models trained on 8$\times$ A800 GPUs. The optimizer parameters are configured with betas set to (0.9, 0.999), momentum set to 0.9, an initial learning rate of $1 \times 10^{-3}$, a weight decay of 0.05, and Exponential Moving Average (EMA).

\noindent \textbf{Results.} Table \ref{tab:imagenet} presents a comparison of our proposed DefMamba models (T, S, B) with various state-of-the-art (SOTA) methods. Specifically, DefMamba-T achieves 78.6\% Top-1 Acc., outperforming CNNs based RegNetY-800M \cite{regnety} and transformer based DeiT-Ti \cite{deit} by 2.3\% and 6.4\%, respectively. Moreover, DefMamba-T outperformes the recently introduced SSMs models, achieving 2.5\%, 2.4\%, and 1.3\% higher performance than ViM-T \cite{vim}, LocalViM-T \cite{localmamba}, and MSVMamba-N \cite{MSV} in terms of parameter and computational complexity. Moreover, it reduces the computational burden by 60\% while achieving a performance improvement of 0.7\% over PlainMamba-L1 \cite{plainmamba}. DefMamba-S achieves 83.5\% Top-1 Acc., surpassing GrootV-T \cite{grootvl} and EfficientVMamba-B \cite{efficientvmamba}. Furthermore, DefMamba-B achieves an accuracy of 84.2\%, exceeding VMamba-S \cite{vmamba} by 0.6\%, demonstrating the effectiveness of our methods.

\begin{table}
    \belowrulesep=0pt
    \aboverulesep=0pt
    \centering
    \resizebox{\linewidth}{!}{\setlength{\arrayrulewidth}{1pt} 
\begin{tabular}{>{\raggedright\arraybackslash}m{3.5cm} |>{\centering\arraybackslash}m{1.2cm} >{\centering\arraybackslash}m{1.2cm} |>{\centering\arraybackslash}m{1.2cm} >{\centering\arraybackslash}m{1.2cm}}
\toprule[2pt]
\multicolumn{5}{c}{\textbf{ADE20K with crop size 512}}  \\
\midrule[1pt]
\textbf{Backbone} & \shortstack{\\[0.5pt] \\ \textbf{mIOU} \\ \textbf{(SS)} \\ [0.5pt]} & \shortstack{\textbf{mIOU} \\ \textbf{(MS)}} & \textbf{\#Param.} & \textbf{\#FLOPs} \\
\midrule[1pt]
ResNet-50 \cite{resnet} & 42.1 & 42.8 & 67M & 953G \\
Swin-T \cite{swin} & 44.5 & 45.8 & 60M & 945G \\
ConvNeXt-T \cite{convnext} & 46.0 & 46.7 & 60M & 939G \\
NAT-T \cite{nat} & 47.1 & 48.4 & 58M & 934G \\
Vim-S \cite{vim} & 44.9 & - & 46M & - \\
LocalVim-S \cite{localmamba} & 46.4 & 47.5 & 58M & 297G \\
VMamba-T \cite{vmamba} & 47.9 & 48.8 & 62M & 949G \\
PlainMamba-L2 \cite{plainmamba} & 46.8 & - & 55M & 285G \\
LocalVMamba-T \cite{localmamba} & 47.9 & 49.1 & 57M & 970G \\
MSVMamba-T \cite{MSV} & 47.6 & 48.5 & 65M & 942G \\
QuadMamba-S \cite{quadmamba} & 47.2 & 48.1 & 62M & 961G \\
GrootV-T \cite{grootvl} & 48.5 & 49.4 & 60M & 941G \\
\rowcolor{gray!20}
\textbf{DefMamba-S} & 48.8 & 49.6 & 65M & 946G \\
\bottomrule[2pt]
\end{tabular}}
    \caption{Semantic segmentation performance on ADE20K val set. The crop size is all set to $512^{2}$. SS and MS denote single-scale and multi-scale testing, respectively.}
    \label{tab:ade20k}
\end{table}


\subsection{Object Detection}

\noindent \textbf{Settings.} We evaluated DefMamba on the MSCOCO 2017 dataset \cite{coco} using the Mask R-CNN framework \cite{maskrcnn} for object detection and instance segmentation tasks. Following prior works \cite{MSV, swin, grootvl}, we utilized backbones pretrained on ImageNet-1K for initialization. We employed standard training strategies, including 1 × (12 epochs) with Multi-Scale (MS) training, to ensure a fair comparison.

\noindent \textbf{Results.} As depicted in Table \ref{tab:coco}, our method outperforms existing methods on most evaluation metrics. Specifically, under 1$\times$ schedule, DefMamba-S achieves 47.5 in box mAP (AP\textsuperscript{b}) and 42.8 in mask mAP (AP\textsuperscript{m}). Our model surpassed ResNet-50 \cite{resnet}, Swin-T \cite{swin} and ConvNeXt-T \cite{convnext}. At the same time, our method improved AP\textsuperscript{m} by 0.6 points compared to LocalVMamba-T \cite{localmamba} and QuadMamba-S \cite{quadmamba}. Furthermore, our method exhibited comparable performance to the previous SOTA method, VMamba-T \cite{vmamba}, while reducing the computational load by 4\%.

\begin{table}
  \centering
  \belowrulesep=0pt
  \aboverulesep=0pt
  \resizebox{\linewidth}{!}{ 
    \setlength{\arrayrulewidth}{1pt} 
\begin{tabular}{c|cccc|cc|c}
\toprule[2pt]
Index & FB-BB  &CB \cite{plainmamba} &LB \cite{localmamba}    &DB  &\#Param.    &\#FLOPs     & Top-1 \\
\midrule[1pt]
(1) & $\surd$  &          &       &    &7.3M  &1.1G  & 76.9 \\
(2) &          &          &       &$\surd$   &8.1M &1.1G & 76.5 \\
(3) &$\surd$   &$\surd$   &       &    &7.5M  &1.1G  &77.3 \\
(4) &$\surd$   &          &$\surd$&    &7.5M  &1.1G &77.1 \\
\rowcolor{gray!20}
(5) & $\surd$  &          &       &$\surd$     &8.3M  &1.2G &\textbf{78.6} \\
\bottomrule[2pt]
\end{tabular}
  }
  \caption{Comparisons with the different scanning branches in the setting of tiny model size. DB means our deformable branch. The best results are shown in \textbf{bold} font.}
  \label{tab:alb1}
\end{table}

\subsection{Semantic Segmentation}

\noindent \textbf{Settings.} To evaluate the semantic segmentation performance of DefMamba, we trained our models using UperNet \cite{upernet} initialized with pre-trained classification weights on ADE20K \cite{ade20k} for 160,000 iterations. We employed the AdamW optimizer \cite{adamw} with a learning rate set at $6 \times 10^{-5}$. Our experiments are primarily conducted using a default input resolution of 512 × 512. Additionally, we incorporated Multi-Scale (MS) testing to assess performance variations.

\noindent \textbf{Results.} The DefMamba-S model demonstrates favorable performance in semantic segmentation compared to various SOTA methods, as presented in Table \ref{tab:ade20k}. DefMamba-S achieves a mIOU of 48.8 in single-scale and 49.6 in multi-scale evaluation. This outperforms ResNet-50 \cite{resnet}, Swin-T \cite{swin}, and ConvNeXt-T \cite{convnext}. Additionally, DefMamba-S exceeds the performance of the recent SSM methods, including GrootV-T \cite{grootvl}, QuadMamba-S \cite{quadmamba} and MSVMamba-T \cite{MSV}. Our method achieves improvements of 0.3 points, 1.6 points, and 1.2 points, respectively, on the single-scale mIoU metric.

\begin{table}
  \centering
  \belowrulesep=0pt
  \aboverulesep=0pt
  \resizebox{\linewidth}{!}{ 
    \setlength{\arrayrulewidth}{1pt} 
\begin{tabular}{c|cccc|cc|c}
\toprule[2pt]
Index & DP & DT & OB & CA & \#Param. & \#FLOPs & Top-1 \\
\midrule[1pt]
(1) &  &  &  &  & 7.5M & 1.1G & 77.0 \\
(2) & $\surd$  &  &  &  & 7.5M & 1.2G & 77.4 \\
(3) &  & $\surd$  &  &  & 7.5M & 1.2G & 77.2 \\
(4) & $\surd$  & $\surd$  &  &  & 7.5M & 1.2G & 77.9 \\
(5) & $\surd$ & $\surd$ & $\surd$ &  & 8.2M & 1.2G & 78.2 \\
\rowcolor{gray!20}
(6) & $\surd$ & $\surd$ & $\surd$ & $\surd$ & 8.3M & 1.2G & \textbf{78.6} \\
\bottomrule[2pt]
\end{tabular}
  }
  \caption{Comparisons with the different components of proposed deformable scanning in tiny model size settings on ImageNet-1k. The best results are shown in \textbf{bold} font.}
  \label{tab:alb2}
\end{table}

\subsection{Ablation Study}
\noindent \textbf{Effect of the Proposed DSSM Structure}. To give the evidence for the effectiveness of the proposed deformable branch, we conducted a series of experiments over different branch settings in Table \ref{tab:alb1}. As shown in Table \ref{tab:alb1}, FB-BB refers to the forward and the backward branches for feature extraction. CB represents continuous scanning \cite{plainmamba} branch. LB is local scanning \cite{localmamba} branch and DB denotes our proposed deformable branch in Figure \ref{fig:overall} (a). By comparing the results in Table \ref{tab:alb1} (1) and (5), we observed that the proposed deformable branch significantly increased the accuracy of the ImageNet dataset by 1.7\% within a reasonable computational budget, to demonstrate the effectiveness of the proposed deformable branch. Moreover, compared with Table \ref{tab:alb1} (3), (4) and (5), our method only increased the computational cost by 0.1G, while significantly improving the accuracy on ImageNet dataset by 1.4\%. This further demonstrates that our deformable scanning approach, as opposed to other fixed scanning methods, is more capable of capturing structural information of objects and enhancing model performance. From Table \ref{tab:alb1} (1) and (2), we noted that incorporating only the proposed deformable branch would lead to a decrease in performance due to the increased spatial token jumps. To achieve more stable training and higher model performance, we adhered to the previous paradigm by combining the FB-BB and DB in our proposed model.

\begin{figure}
    \centering
    \includegraphics[width=\columnwidth]{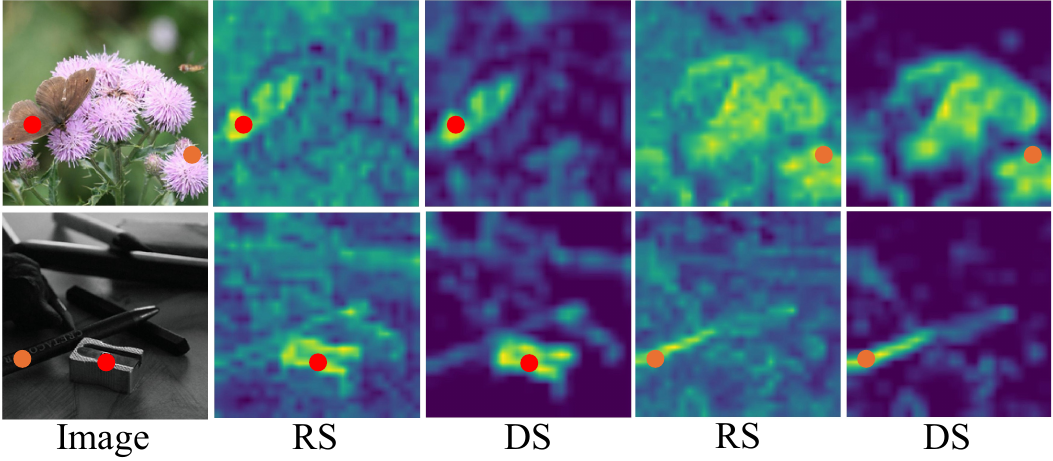}
    \caption{Visualization of activation maps in the specific position. The position is marker by red and orange point. RS stands for raster scanning, DS stands for our deformable scanning.}
    \label{fig:affi}
\end{figure}

\noindent \textbf{Effect of the Deformable Scanning Components}. To thoroughly evaluate the contributions of each component in our proposed deformable scanning method, we conducted an ablation study in Table \ref{tab:alb2}. As shown in Table \ref{tab:alb2}, DP, DT, OB and CA represent the components of the deformable scanning in Figure \ref{fig:overall} (b), respectively. DP (Deformable Points) involves the process of generating deformable points, which includes initializing reference points, an offsets network and a bilinear interpolation. DT (Deformable Tokens) refers to the process of dynamically changing the token order based on predicted offsets, which includes initializing token index and deformable token index, as well as an offsets network. OB represents the operation of generating offset bias. CA denotes channel attention operation in the offset network. By comparing the results in Table \ref{tab:alb2} (1), (2) and (3), we observed that adding either the DP operation or the DT operation boosting the baseline (1) performance by 0.2-0.4\%, with only a minor increases in computational costs (0.1G), demonstrated that the effectiveness of both DP and DT. Moreover, when combining both DP and DT operations (4), the performance further improves by 1\% compared to the baseline, as shown in Table \ref{tab:alb2} (1) and (4). These results strongly prove that whether adding DT or DP individually, or combining both, the model can better learn the structure of images and facilitate the awareness of changes in object details, thus improving performance. Furthermore, we run ablation experiments to confirm the effectiveness of OB and CA by comparing Table \ref{tab:alb2} (4) and (5), and Table \ref{tab:alb2} (5) and (6), respectively. The experiments validate the effectiveness of our methods.

\noindent \textbf{Visualization Results}. To better demonstrate the superiority of our deformable scanning strategy, we present the activation maps of images at different positions for various methods in Figure \ref{fig:affi}, marked clearly by red and orange dots in the image. Specifically, we visualized the activation map corresponding to the final layer of the second stage using the method outlined in \cite{vmamba}. As illustrated in the activation map of the pen in the last line of Figure \ref{fig:affi}, our method demonstrates an enhanced ability to focus on the structural and shape information of objects, even when dealing with complex scenes that contain multiple overlapping objects. This capability allows for more precise recognition and segmentation, further highlighting the effectiveness of our approach in capturing essential details.

\begin{figure}
    \centering
    \includegraphics[width=\columnwidth]{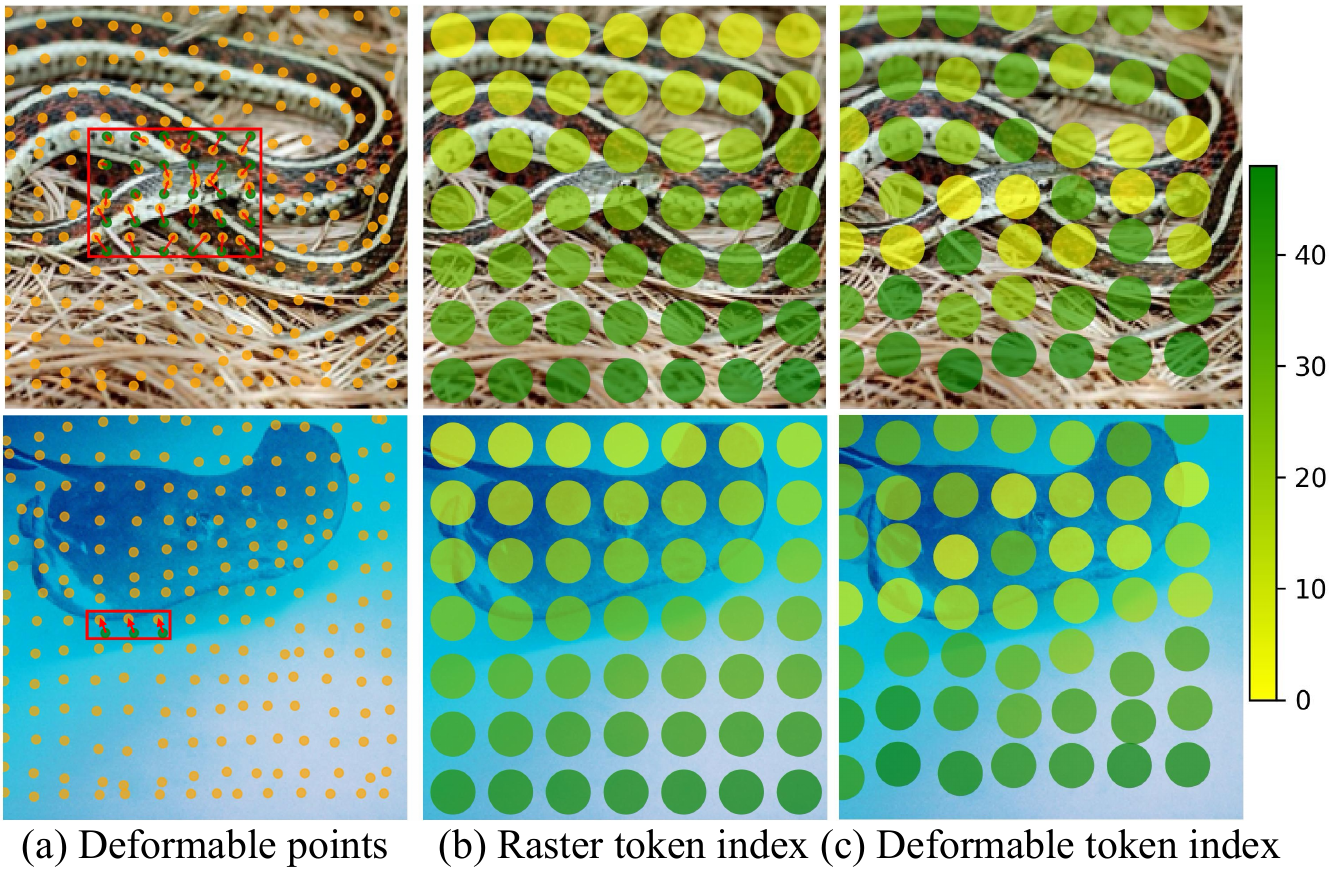}
    \caption{Visualization of deformable points and deformable token index. In (a), the orange dots represent deformable points, the green dots represent reference points, and the red arrows represent the offset path of the points. In (b) and (c), the gradient from yellow to green represents the scanning path, with the yellow dots being scanned first and the green dots being scanned later.}
    \label{fig:re}
\end{figure}

We also visualized the deformable points and deformable token index to intuitively demonstrate the performance of our method, as shown in the Figure \ref{fig:re}. In the red box of Figure \ref{fig:re} (a), we can observe that some focus points outside the object shift towards the object, Such movements allow the network to attend to more object information. Compared with Figure \ref{fig:re} (b), our method (c) adjusts the scanning order to emphasize important tokens. For example, as shown in the first row of the Figure \ref{fig:re}, the token corresponding to the snake's head is shifted from the middle position in a raster scan to being the first position in our method. Such offsets are beneficial for the network to learn relevant features. 

\section{Limitation}
\label{sec:weak}

\begin{figure}
    \centering
    \includegraphics[width=\columnwidth]{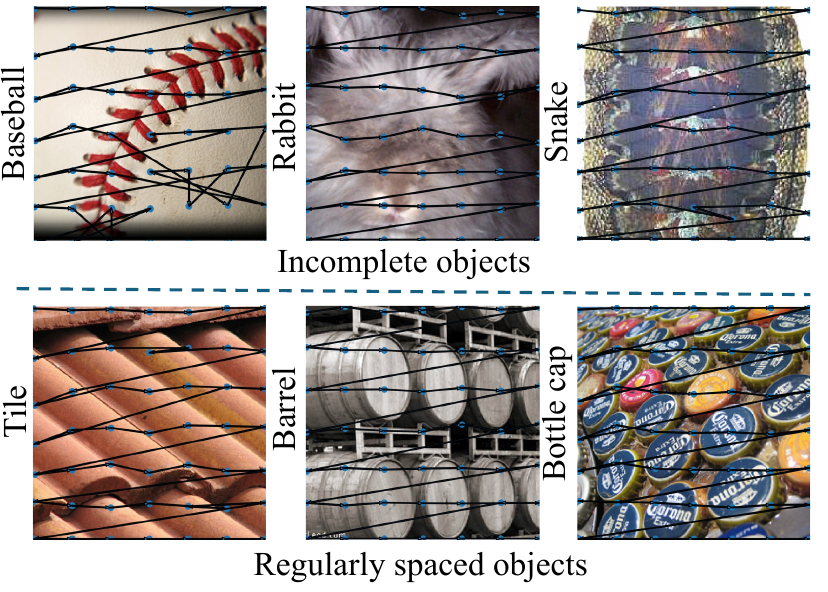}
    \caption{Failure cases. The blue dots represent deformable points, while the black arrows indicate the scanning order after deformation.}
    \label{fig:limi}
\end{figure}

Despite the strong results of our approach, there is still a limitation. In cases where the image contains incomplete object structures or multiple objects arranged in a regular pattern, the deformable scanning strategy may be less effective. As illustrated in Figure \ref{fig:limi}, when the image shows only a portion of a baseball, the deformable mechanism does not capture complete structural information, resulting in offsets that are too small and converge towards the predefined scanning method. Meanwhile, when multiple objects are arranged according to a certain rule, the information variation between adjacent tokens is minimal. Results in the model remaining in an indolent learning process.

\section{Conclusion}
\label{sec:conc}
In this work, we made efforts to tackle the challenge arising from existing Mamba-based methods, which rely on fixed scanning techniques to extract features. These methods do not fully leverage the spatial structure information inherent in images. To overcome this limitation, we proposed a novel foundation model named DefMamba, which is grounded in DS. This innovative model aims to enhance the capability of the network to learn and represent complex image structures, as well as to detect subtle changes in object details. The DS principally comprises two key operations: the shifting of focus points and the alteration of the scanning order. The first operation effectively repositions reference points toward significant regions of interest, thereby enhancing the model's sensitivity to variations in object details. The second operation modifies the scanning order to create a structure-aware sequence that is better aligned with the underlying object structure based on the input features. Extensive experimental evaluations on benchmark datasets, including ImageNet, COCO, and ADE20K, robustly demonstrate that our proposed method outperforms existing SSMs and remains competitive when compared to both CNNs and Transformer-based approaches.

\section*{Acknowledgement}
This work was supported by the National Natural Science Foundation of China (62172070, 62372080, 62376050, and U22B2052), and the Natural Science Foundation of Liaoning Province (2024-MSBA-24).

{
    \small
    \bibliographystyle{ieeenat_fullname}
    \bibliography{main}
}

\clearpage
\renewcommand{\thesection}{\Alph{section}}
\renewcommand\thefigure{\Alph{section}\arabic{figure}}
\renewcommand\thetable{\Alph{section}\arabic{table}}
\setcounter{page}{1}
\setcounter{section}{0}
\setcounter{figure}{0}
\setcounter{table}{0}

\maketitlesupplementary

\section{Model Detailed Structure}
\label{sec:detail}
In Table \ref{tab:detail}, we present the detailed architecture of our model variants, including the Tiny, Small, and Base versions, each with varying numbers of channels and blocks. During the experimental process, considering the channel redundancy mentioned in previous methods \cite{groupmamba, MSV} and the experimental findings in \cite{vmamba}, we set the SSM\_RATIO to 1. Following prior work \cite{grootvl}, we employed two $3\times3$ convolutions with a stride of 2 and padding of 1 in the patch embedding layer, interspersed with LN and GELU activation. For the downsampling layer, we utilized a $3\times3$ convolution with a stride of 2 and padding of 1, followed by LN.

\begin{table}[ht]
\centering
\resizebox{\linewidth}{!}{ 
\begin{tabular}{>{\centering\arraybackslash}m{1.2cm} >{\centering\arraybackslash}m{2.7cm} >{\centering\arraybackslash}m{2.7cm} >{\centering\arraybackslash}m{2.7cm}}
\toprule[2pt]
\textbf{Stage} & \textbf{DefMamba-T} & \textbf{DefMamba-S} & \textbf{DefMamba-B} \\ 
\midrule
(224$\times$224) & \multicolumn{3}{c}{Patch embedding} \\
\midrule
\shortstack{\textbf{Stage 1} \\ (56$\times$56)} & 
\shortstack{$N_1 = 2, \, C = 48$ \\ $K$: $9$} &
\shortstack{$N_1 = 2, \, C = 96$ \\ $K$: $9$} &
\shortstack{$N_1 = 2, \, C = 96$ \\ $K$: $9$} \\ 
\midrule
\shortstack{\textbf{Stage 2} \\ (28$\times$28)} & 
\shortstack{$N_2 = 2, \, C = 96$ \\ $K$: $7$} &
\shortstack{$N_2 = 2, \, C = 192$ \\ $K$: $7$} &
\shortstack{$N_2 = 3, \, C = 192$ \\ $K$: $7$} \\ 
\midrule
\shortstack{\textbf{Stage 3} \\ (14$\times$14)} & 
\shortstack{$N_3 = 5, \, C = 192$ \\ $K$: $5$} &
\shortstack{$N_3 = 6, \, C = 384$ \\ $K$: $5$} &
\shortstack{$N_3 = 16, \, C = 384$ \\ $K$: $5$} \\ 
\midrule
\shortstack{\textbf{Stage 4} \\ (7$\times$7)} & 
\shortstack{$N_4 = 2, \, C = 384$ \\ $K$: $3$} &
\shortstack{$N_4 = 2, \, C = 768$ \\ $K$: $3$} &
\shortstack{$N_4 = 2, \, C = 768$ \\ $K$: $3$} \\ 
\midrule
(1$\times$1) & \multicolumn{3}{c}{Average pool, 1000-d FC, Softmax} \\
\midrule
\textbf{\#Param.} & 8M & 32M & 51M \\
\midrule
\textbf{\#FLOPs} & 1.2G & 4.8G & 8.5G \\
\bottomrule[2pt]
\end{tabular}
}
\caption{Model architecture specifications. $N_i$ represents the number of DM blocks in the i-th stage. $C$ represents the number of channels. $K$ represents the kernel size of the DWConv in the Offset network. FC represents the fully connected layer.}
\label{tab:detail}
\end{table}

\section{More variants of object detection}
To further validate the performance of our designed DefMamba variants on the object detection and instance segmentation task, we conducted experiments based on DefMamba-B. As shown in the Table \ref{tab:cocob}.

\begin{table}[ht]
    \belowrulesep=0pt
    \aboverulesep=0pt
    \centering
    \resizebox{\linewidth}{!}{
    \setlength{\arrayrulewidth}{1pt} 
    \begin{tabular}{l|c|ccc|ccc}
    \toprule[2pt]
    \multicolumn{7}{c}{\textbf{Mask R-CNN 1$\times$ schedule}} \\
    \midrule[1pt]
    \textbf{Backbone} & \textbf{\#FLOPs} & \textbf{AP\textsuperscript{b}} & \textbf{AP\textsuperscript{b}\textsubscript{50}} & \textbf{AP\textsuperscript{b}\textsubscript{75}} & \textbf{AP\textsuperscript{m}} & \textbf{AP\textsuperscript{m}\textsubscript{50}} & \textbf{AP\textsuperscript{m}\textsubscript{75}} \\
    \midrule[1pt]
    ResNet-101 \cite{resnet} & 336G & 38.2 & 58.8 & 41.4 & 34.7 & 55.7 & 37.2 \\
    Swin-S \cite{swin} & 354G & 44.8 & 66.6 & 48.9 & 40.9 & 63.4 & 44.2  \\
    ConvNeXt-S \cite{convnext} & 348G & 45.4 & 67.9 & 50.0 & 41.8 & 65.2 & 45.1  \\
    VMamba-S \cite{vmamba} & 400G & 48.2 & 69.7 & 52.5 & 43.0 & 66.6 & 46.4  \\
    LocalVMamba-S \cite{localmamba} & 414G & 48.4 & 69.9 & 52.7 & 43.2 & 66.7 & 46.5  \\
    GrootV-S \cite{grootvl} & 341G & 48.6 & 70.3 & 53.5 & 43.6 & 67.5 & 47.1 \\
    \rowcolor{gray!20}
    \textbf{DefMamba-B} & 349G & 48.7 & 70.5 & 53.8 & 43.7 & 67.7 & 47.0 \\
    \bottomrule[2pt]
    \end{tabular}}
    \caption{Performance of Object Detection and Instance Segmentation Based on DefMamba-B on COCO. AP\textsuperscript{b} and AP\textsuperscript{m} indicate the mean Average Precision (mAP) of detection and segmentation.}
    \label{tab:cocob}
\end{table}

As shown in the table above, our method still outperforms previous methods at the base scale.

\section{More variants of semantic segmentation}
To further validate the performance of our designed DefMamba variants on the semantic segmentation task, we conducted experiments based on DefMamba-B. As shown in the Table \ref{tab:ade20kb}.

\begin{table}[ht]
    \belowrulesep=0pt
    \aboverulesep=0pt
    \centering
    \resizebox{\linewidth}{!}{
    \setlength{\arrayrulewidth}{1pt} 
    \begin{tabular}{>{\raggedright\arraybackslash}m{3.5cm} |>{\centering\arraybackslash}m{1.2cm} >{\centering\arraybackslash}m{1.2cm} |>{\centering\arraybackslash}m{1.2cm} >{\centering\arraybackslash}m{1.2cm}}
    \toprule[2pt]
    \multicolumn{5}{c}{\textbf{ADE20K with crop size 512}}  \\
    \midrule[1pt]
    \textbf{Backbone} & \shortstack{\\[0.5pt] \\ \textbf{mIOU} \\ \textbf{(SS)} \\ [0.5pt]} & \shortstack{\textbf{mIOU} \\ \textbf{(MS)}} & \textbf{\#Param.} & \textbf{\#FLOPs} \\
    \midrule[1pt]
    ResNet-101 \cite{resnet} & 42.9 & 44.0 & 85M & 1030G \\
    Swin-S \cite{swin} & 47.6 & 49.5 & 81M & 1039G \\
    ConvNeXt-S \cite{convnext} & 46.0 & 46.7 & 60M & 939G \\
    VMamba-S \cite{vmamba} & 49.5 & 50.5 & 76M & 1081G \\
    PlainMamba-L3 \cite{plainmamba} & 49.1 & - & 81M & 419G \\
    LocalVMamba-S \cite{localmamba} & 50.0 & 51.0 & 81M & 1095G \\
    QuadMamba-B \cite{quadmamba} & 49.7 & 50.8 & 82M & 1042G \\
    GrootV-S \cite{grootvl} & 50.7 & 51.7 & - & 1019G \\
    \rowcolor{gray!20}
    \textbf{DefMamba-B} & 50.8 & 51.7 & 84M & 1024G \\
    \bottomrule[2pt]
    \end{tabular}
    }
    \caption{Performance of Semantic Segmentation Based on DefMamba-B on ADE20K. SS and MS denote single-scale and multi-scale testing, respectively.}
    \label{tab:ade20kb}
\end{table}

As shown in the table above, our method is comparable to previous methods at the base scale.

\end{document}